\documentclass[runningheads]{llncs}
\usepackage[T1]{fontenc}
\IfFileExists{lmodern.sty}{\usepackage{lmodern}}{}
\usepackage[utf8]{inputenc}
\usepackage{graphicx}
\usepackage{booktabs}
\usepackage{amsmath}
\usepackage{amssymb}
\usepackage{pgfplots}
\pgfplotsset{compat=1.18}
\usepackage{marvosym}
\usepackage[htt]{hyphenat}
\usepackage{xurl}
\usepackage[a4paper,textwidth=14.6cm,textheight=21.6cm,centering]{geometry}
\usepackage[hidelinks]{hyperref}
\let\citep\cite

\title{Can Agentic Systems Close the Gap in Radiology with Production AI Tools?}
\titlerunning{Can Agentic Systems Close the Gap in Radiology?}
\author{Aubin Charley\inst{1} \and
Julien Khlaut\inst{1} \and
Charles Corbi\`ere\inst{1} \and
Pierre Manceron\inst{1} \and
Corentin Dancette\inst{1} \and
L\'eo Machado\inst{1,2}(\Letter)}
\authorrunning{A. Charley et al.}
\institute{Raidium, Paris, France \and
Department of Radiology, Fondation Ophtalmologique Adolphe de Rothschild, Paris, France\\
\email{dr.machado.leo@gmail.com}}

\begin{document}

\maketitle

\begin{abstract}
Radiology AI has developed along two paths: specialized models that perform well on narrow tasks but cover few clinical findings, and generalist models that address a broader range of tasks but remain prone to hallucination and inaccurate measurements. Combining the two through agentic systems offers a route towards comprehensive radiology reporting, but published systems are tied to one modality or body part. We describe a radiology agentic framework that standardizes how an orchestrator calls modular tools to produce a structured report, and use it to measure what such a system can do with the tools from FDA-cleared radiology AI devices. Report generation is simulated on a large multi-modality report corpus, with the orchestrator given either a catalog of finding classes consolidated from the cleared devices, each available on the modalities it is cleared for, or an unlimited catalog in which any class can be requested, and planning either from the clinical indication or from the target report. Tool outputs are simulated from the reference report, so the comparison isolates tool availability and orchestration from tool error. With the cleared catalog and the indication alone, the orchestrator recovers under a third of the clinically actionable findings; recovery improves substantially with the unlimited catalog, and further when the plan is written from the target report. Attributing every missed finding to a cause shows that most are classes no cleared device offers for the study's modality, and that a third are classes or tools that were offered and not called. Adding the most frequently missing tools closes most of the gap to the unlimited catalog on held-out reports, whereas a smaller orchestrator with the unlimited catalog gives up the whole gain. Tool availability and effective orchestration are thus complementary constraints on comprehensive AI-assisted radiology reporting.
\end{abstract}

\section{Introduction}

A radiology examination starts with the reading of the images (radiographs, CT, MRI, ultrasound) and ends in a report: a text in which the radiologist states what the images show, what is normal and what it means for the patient, condensing hundreds of observations, many of them made without ever being put into words. Over the past decade, AI in radiology has taken the form of single-purpose devices, each cleared by a regulator for one task on one modality; the US Food and Drug Administration (FDA) lists more AI-enabled devices for radiology than for any other specialty~\citep{fda-aiml}. A report, however, is not one task. Reading a chest CT for cancer staging means surveying the lungs, the pleura, the mediastinum, the vessels, the bones and the upper abdomen, measuring what is found, comparing it with the previous study and weighing the findings together; a tool that detects lung nodules covers one step of that sequence. The question for AI in radiology has therefore moved from whether one finding can be detected to whether a system can assemble many such tools into the reading of a whole study, and this paper measures how far that assembly can go with the tools cleared today.

Two kinds of AI tools coexist in radiology. Specialized models (lung nodule detection, liver segmentation, calcium scoring) are accurate within their scope but cannot be composed to cover the thousands of findings that appear in reports, many of which are never named explicitly and emerge from the radiologist's reading. Vision-language models (VLMs) produce fluent reports but omit findings, hallucinate and measure imprecisely~\citep{ct2rep,wang2026hallucxr}; in 3D CT report generation they also fall back on stereotyped text, a failure described as template collapse~\citep{templatecollapse}. Agentic systems, in which a large language model (LLM) orchestrator plans and calls tools, combine the two and have outperformed either alone in several settings~\citep{medagent3d,radiologistcopilot,medrax,ctflow,petctagent,tripathi,khandelwal,bluethgen2025}. The published systems are, however, built for one modality or one anatomy with a fixed tool set: chest radiographs~\citep{medrax}, CT~\citep{ctflow}, PET/CT~\citep{petctagent}, or a radiotherapy pathway~\citep{jiang2026radonc}.

Our design borrows from the Model Context Protocol (MCP)~\citep{mcp}, an open standard from the software industry under which a language model discovers and calls external tools and data sources through one common interface, whoever built them, much as a USB port lets any device connect to a computer. On the orchestrator side it follows the ReAct pattern~\citep{react}, in which the model alternates between reasoning about what it has seen so far and acting by calling a tool, each result feeding the next decision, so that a report is produced by a sequence of such cycles over the study rather than by a single generation. We propose a radiology agentic framework with the same shape, a shared interface: a vocabulary of resources (patient context, prior studies, templates, guidelines) and tools (viewer interaction, detection, segmentation, quantification, longitudinal comparison, reporting) that any compliant orchestrator can compose, and that can run against different deployed tool catalogs.

We use the framework to ask what an agentic system can do with the tools cleared for clinical use today. We compare an orchestrator given a tool-infinite catalog, where it can request any plausible tool and finding class, with the same orchestrator restricted to a catalog distilled from the FDA list of AI-enabled devices for radiology: 1{,}164 cleared devices, consolidated into 1{,}989 finding classes, each available on the modalities it is cleared for~\citep{fda-aiml}. Tool outputs are simulated from the reference report in both cases, so the comparison measures tool availability and orchestration rather than tool accuracy. The distance between the two catalogs bounds what new tools could add; the distance between the tool-infinite catalog and the reference report bounds the agentic design itself. We then attribute every finding the cleared-catalog orchestrator misses to a cause, test the simplest fix, adding the most frequently missing tools, on held-out reports, and measure how much of the tool-infinite result depends on the orchestrator model. The evaluation uses PARROT, a corpus of fictional reports across modalities and anatomical regions~\citep{parrot}.

\section{Materials and Methods}

\subsection{The Agentic Framework}
\label{sec:architecture}

\subsubsection{Architecture.}
\begin{figure}[!htbp]\centering
\includegraphics[width=\linewidth]{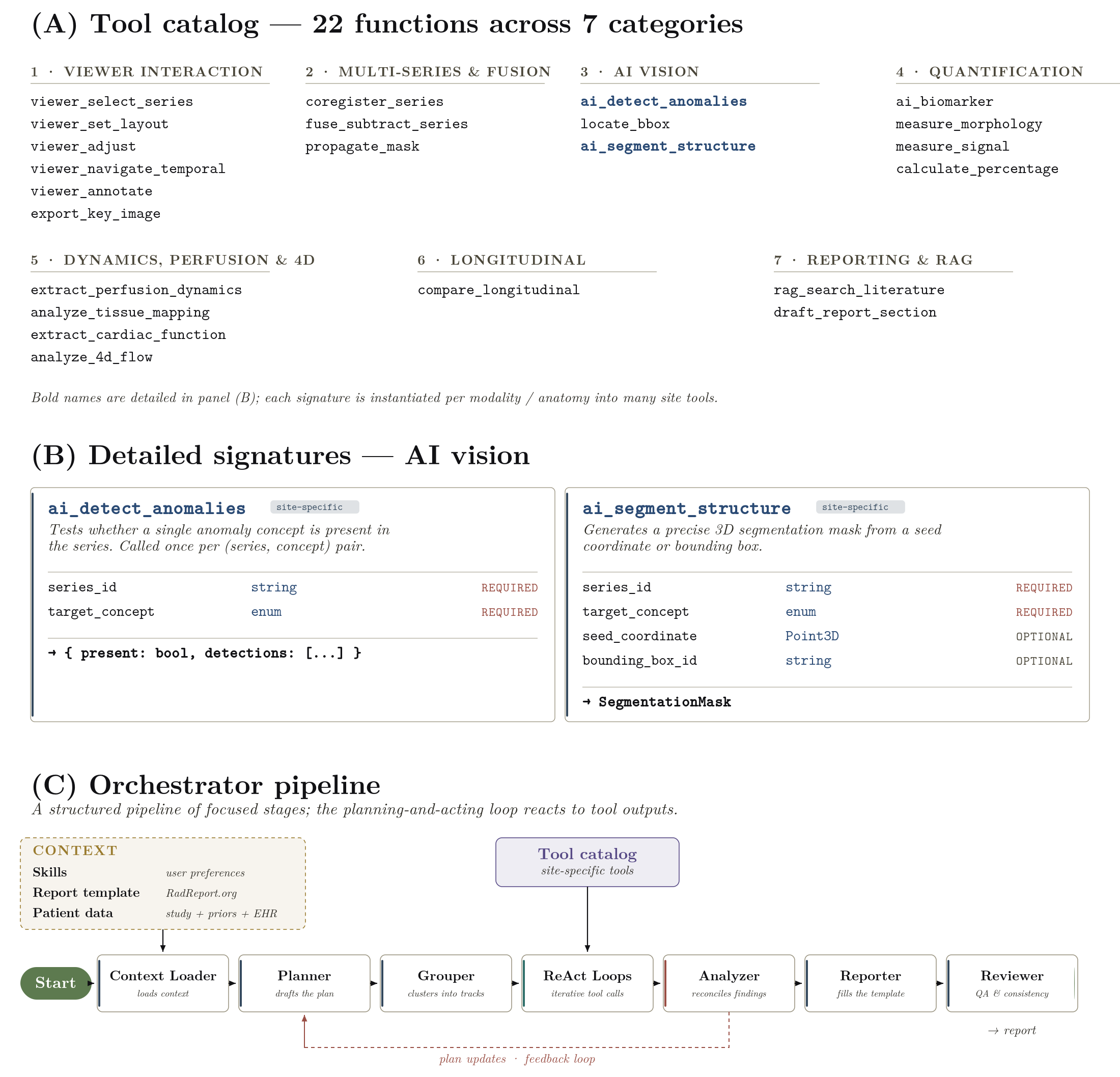}
\caption{Overview of the framework. (\textbf{A}) Function catalog: 22 tool functions across 7 categories; functions are instantiated from the site-specific tool catalog. (\textbf{B}) Detailed signatures for two representative AI-vision functions, with required and optional parameters and the structured return type. (\textbf{C}) Orchestrator pipeline: clinical context (skills, RadReport.org template, patient data) is loaded by the Context Loader and feeds the Planner, which dispatches steps via the Grouper to parallel ReAct Loops that call tools from the available tool catalog; the Analyzer synthesises across tracks and may issue plan updates back to the Planner (feedback loop, dashed) before handing off to the Reporter and the Reviewer.}
\label{fig:architecture}
\end{figure}

The framework has three layers (Figure~\ref{fig:architecture}): \textbf{resources} (patient context, prior studies, a whole-body anatomical reference frame from a total-body segmentation, clinical guidelines, RadReport.org templates, the catalog of available tools, and the study under interpretation); \textbf{tools} in seven categories (viewer interaction, multi-series operations, AI vision, quantification, dynamic analysis, longitudinal comparison, clinical intelligence); and an \textbf{orchestrator} that plans, calls tools and writes a structured report. When a tool is unavailable the orchestrator can skip the step or flag it in the report.

\subsubsection{Orchestrator Pipeline.}
The orchestrator is a pipeline of stages with defined inputs and outputs rather than a single LLM call; together they implement the planning-and-acting loop that earlier systems describe monolithically~\citep{medagent3d,radiologistcopilot,medrax,ctflow,petctagent,react}. The \textbf{Context Loader} reads the patient record, retrieves prior studies and resolves the tools available. The \textbf{Planner} turns the clinical indication and a retrieved template into a flat, ordered list of analysis steps (detection, measurement, comparison) followed by two drafting steps; each drafting step carries a \emph{coverage manifest} naming, in reporting order, the anatomical regions and observation categories the report must cover. The \textbf{Grouper} partitions the steps into tracks that can run in parallel: steps on the same lesion or series (detect, segment, measure) stay in one sequential track, steps on independent series run concurrently; it adds no steps of its own. One \textbf{ReAct loop} runs per track, executing steps one at a time, distilling tool results into findings, reacting to detection or longitudinal results in the middle of the loop, and stopping when its steps are exhausted or a budget is reached. The \textbf{Analyzer} merges the tracks, reconciles discrepancies and decides whether to proceed, re-group, or replan. The \textbf{Reporter} drafts the report in the template's sections from the per-track conclusions and measurements. The \textbf{Reviewer} scores the draft against the findings the Reporter used and triggers one in-place rewrite when the score falls below a threshold.

Restricting the tool catalog changes only which calls the ReAct loops can make. The Planner, Analyzer, Reporter and Reviewer see the same context and the same prompts in every mode, so the differences reported in Section~\ref{sec:results} are due to tool availability.

\subsection{Dataset}

We evaluate on \textbf{PARROT}~\citep{parrot}, an open multilingual collection of 2{,}658 fictional radiology reports written by 76 authors from 21 countries, covering CT (36\%), MRI (23\%), radiography (19\%) and ultrasound (17\%) and the chest, abdomen, head and pelvis. We keep English-language reports that include a clinical indication. The reports for which all four evaluation modes produced a parseable output number 1{,}059. For the catalog-augmentation experiment (Section~\ref{sec:gap-analysis}) they are split once into 659 derivation and 400 test reports.

\subsection{Reporting Templates}

For each report we retrieve a RadReport.org template~\citep{radreport} matching its modality, anatomical region and indication. The orchestrator must follow the template's section structure, which removes layout differences between generated and reference reports and lets the metric focus on content.

\subsection{Cleared-Tool Catalog Construction}
\label{sec:fda-catalog}

The cleared-tool catalog is seeded from the FDA list of AI-enabled medical devices~\citep{fda-aiml}, restricted to the Radiology lead panel: 1{,}164 devices (1{,}146 510(k), 9 De Novo and 9 PMA decisions) across 40 product codes, decided between June 1998 and March 2026. For each device we downloaded the public 510(k) or De Novo decision summary from \texttt{accessdata.fda.gov}, which describes the intended use, the input data and the outputs the device produces.

Classes are extracted from these summaries in two LLM passes. An extractor (gpt-5.4-mini) reads each summary and lists the classes the device outputs, each with a task (detection, segmentation or biomarker), a body system and the modalities it applies to; a second pass re-checks every proposed class against the summary, dropping 415 entries that did not name an output class and renaming 379, which leaves 2{,}901 classes. A consolidation model (gpt-5.6-sol) then groups classes that name the same imaging concept, one body system at a time, and keeps one name per group (\texttt{ablation\_zones} $\rightarrow$ \texttt{ablation\_zone}, \texttt{diffusion\_anisotropy} $\rightarrow$ \texttt{fractional\_anisotropy}, \texttt{lvesv\_a4c} $\rightarrow$ \texttt{end\_systolic\_volume}); five rounds produce 912 merges (660 in the first round) and a final catalog of 1{,}989 classes, each with its body system, task, modalities and the cleared devices that support it.

\begin{sloppypar}
Each task becomes the class list (enum) of one tool function of the framework (Figure~\ref{fig:architecture}A): \texttt{ai\_detect\_anomalies} receives the 509 detection classes, \texttt{ai\_segment\_structure} and \texttt{compare\_longitudinal} share the 571 segmentation classes, and \texttt{ai\_biomarker} receives the 909 biomarker classes. The 18 other tool functions (viewer interaction, measurement, registration, perfusion, cardiac function) take no class argument and are unchanged. In the cleared-catalog modes the orchestrator can only request classes from this list, filtered to the modality of the study: the detection enum offered on a CT study holds the 260 detection classes cleared for CT, and 171, 149 and 137 classes are offered on MR, radiography and ultrasound.
\end{sloppypar}

\begin{figure}[!htbp]
\centering
\begin{minipage}[c]{8.8cm}
\centering
\begin{tikzpicture}
\begin{axis}[
  xbar stacked, scale only axis, width=5.6cm, height=9.6cm, bar width=5.8pt,
  xmin=0, xmax=430,
  symbolic y coords={Retina,Nuclear medicine,Hepatobiliary,Body composition,Fetal,Abdomen,Dental,Scoring / reporting,Pelvis,Image processing,Head and neck,Breast,Spine,Oncology,Vascular,Thorax,Musculoskeletal,Brain,Cardiac,Derived biomarkers},
  ytick=data, enlarge y limits=0.035,
  y tick label style={font=\scriptsize}, x tick label style={font=\scriptsize},
  xlabel={classes}, xlabel style={font=\scriptsize},
  axis line style={gray!60}, tick style={gray!60},
  legend style={at={(0.97,0.04)}, anchor=south east, font=\scriptsize, draw=none, fill=none}, legend cell align=left,
  title={\textbf{A}\quad Classes by body system and task}, title style={font=\scriptsize, at={(0,1)}, anchor=south west},
]
\addplot[fill=blue!60!black, draw=none] coordinates {(10,Retina) (3,Nuclear medicine) (7,Hepatobiliary) (3,Body composition) (6,Fetal) (22,Abdomen) (22,Dental) (13,Scoring / reporting) (8,Pelvis) (11,Image processing) (6,Head and neck) (23,Breast) (18,Spine) (33,Oncology) (38,Vascular) (58,Thorax) (20,Musculoskeletal) (117,Brain) (86,Cardiac) (5,Derived biomarkers)};
\addplot[fill=orange!85!black, draw=none] coordinates {(2,Retina) (0,Nuclear medicine) (5,Hepatobiliary) (9,Body composition) (3,Fetal) (16,Abdomen) (23,Dental) (2,Scoring / reporting) (38,Pelvis) (1,Image processing) (60,Head and neck) (24,Breast) (39,Spine) (49,Oncology) (40,Vascular) (33,Thorax) (86,Musculoskeletal) (81,Brain) (60,Cardiac) (0,Derived biomarkers)};
\addplot[fill=gray!55, draw=none] coordinates {(0,Retina) (14,Nuclear medicine) (6,Hepatobiliary) (13,Body composition) (18,Fetal) (2,Abdomen) (1,Dental) (37,Scoring / reporting) (11,Pelvis) (47,Image processing) (0,Head and neck) (27,Breast) (20,Spine) (16,Oncology) (38,Vascular) (45,Thorax) (40,Musculoskeletal) (25,Brain) (168,Cardiac) (381,Derived biomarkers)};
\legend{Detection (509), Segmentation (571), Biomarker (909)}
\end{axis}
\end{tikzpicture}
\end{minipage}\hfill
\begin{minipage}[c]{5.6cm}
\centering
\begin{tikzpicture}
\begin{axis}[
  ybar, scale only axis, width=3.9cm, height=4.1cm, bar width=9pt,
  ymin=0, ymax=1250, xmin=0.4, xmax=5.6,
  xtick={1,2,3,4,5}, xticklabels={CT,MRI,US,RX,PET},
  ytick={0,400,800,1200},
  x tick label style={font=\scriptsize}, y tick label style={font=\scriptsize},
  nodes near coords, nodes near coords style={font=\tiny},
  axis line style={gray!60}, tick style={gray!60},
  title={\textbf{B}\quad Classes per modality}, title style={font=\scriptsize, at={(0,1)}, anchor=south west, align=left, text width=4.4cm},
]
\addplot[fill=blue!60!black, draw=none] coordinates {(1,1048) (2,863) (3,519) (4,465) (5,271)};
\end{axis}
\end{tikzpicture}

\vspace{2mm}
\begin{tikzpicture}
\begin{axis}[
  ybar, scale only axis, width=3.9cm, height=4.1cm, bar width=9pt,
  ymin=0, ymax=1400, xmin=0.4, xmax=5.6,
  xtick={1,2,3,4,5}, xticklabels={1,2,3--4,5--9,$\geq$10},
  ytick={0,400,800,1200},
  x tick label style={font=\scriptsize}, y tick label style={font=\scriptsize},
  xlabel={cleared devices behind the class}, xlabel style={font=\scriptsize},
  nodes near coords, nodes near coords style={font=\tiny},
  axis line style={gray!60}, tick style={gray!60},
  title={\textbf{C}\quad Devices per class}, title style={font=\scriptsize, at={(0,1)}, anchor=south west, align=left, text width=4.4cm},
]
\addplot[fill=gray!55, draw=none] coordinates {(1,1165) (2,321) (3,257) (4,163) (5,83)};
\end{axis}
\end{tikzpicture}
\end{minipage}
\caption{The cleared-tool catalog: 1{,}989 finding classes consolidated from the 1{,}164 FDA-cleared radiology AI devices (decisions from June 1998 to March 2026). (\textbf{A}) Classes per body system, split by the tool that exposes them; ``derived biomarkers'' are measurements not tied to one organ. (\textbf{B}) Classes available per modality; a class can cover several. (\textbf{C}) Number of cleared devices behind each class: 59\% of the classes come from a single device.}
\label{fig:cleared}
\end{figure}
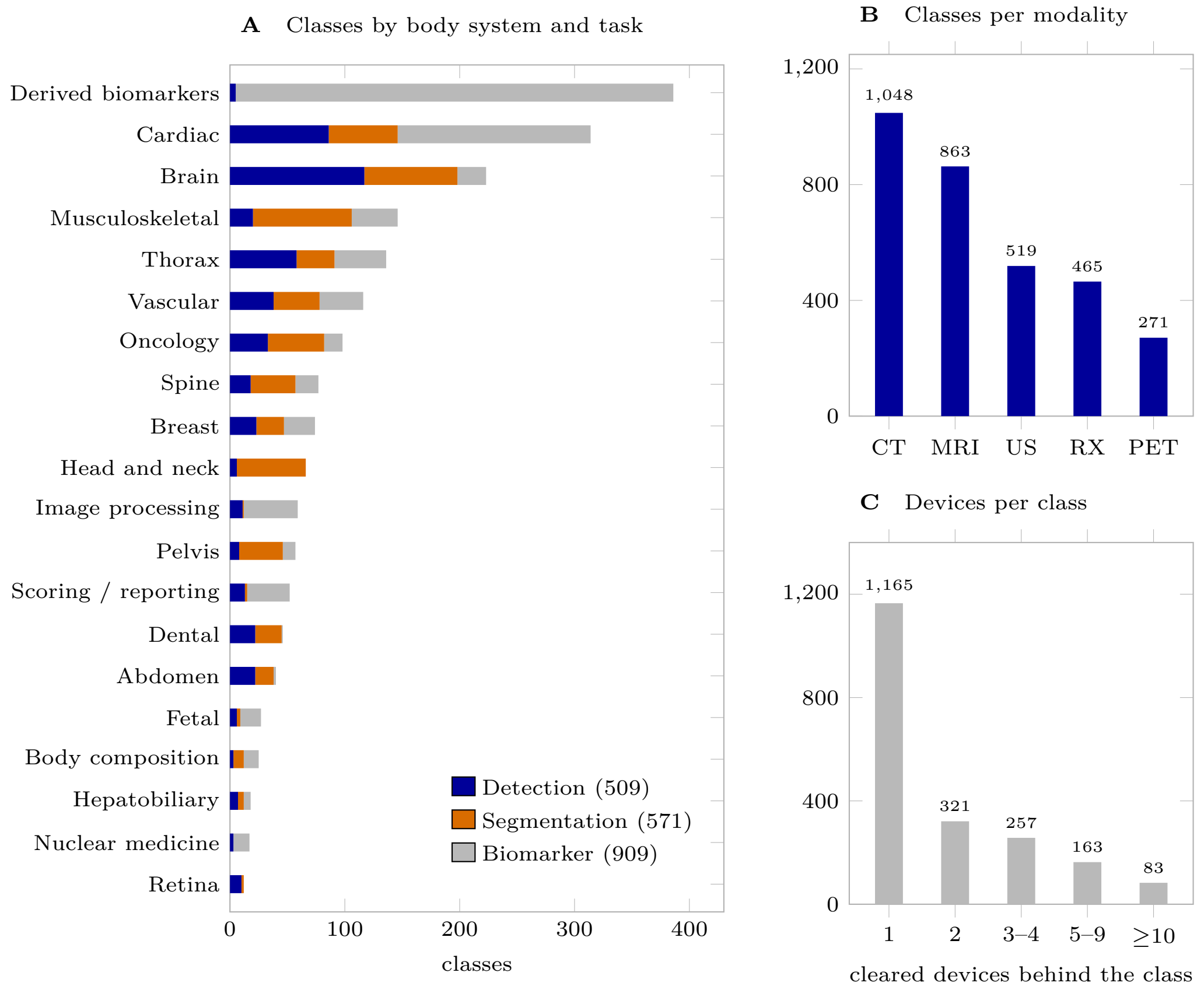

Figure~\ref{fig:cleared} summarises the catalog.

\subsection{Evaluation Modes}
\label{sec:modes}

We vary two factors:

\begin{description}
  \item[Tool catalog.] \textbf{Tool-Infinite}: the orchestrator can request any plausible tool with any finding class for the sub-questions it identifies; tool outputs are simulated from the reference report. This is the upper bound on tool availability. \textbf{Cleared}: the detection, segmentation and biomarker tools accept only the 1{,}989 classes of Section~\ref{sec:fda-catalog}, each on the modalities it is cleared for; outputs are again simulated from the reference, so catalog breadth, not algorithmic error, is the only difference.
  \item[Information given to the Planner.] \textbf{+Report} (oracle): the Planner reads the target report when writing its plan. This controlled leakage measures how well the rest of the pipeline executes a plan written with knowledge of the answer; we report it as a topline, not as a realistic setting. \textbf{+Clinical Indication} (realistic): the Planner sees only the indication and the template.
\end{description}

Crossing the two factors gives four modes, the \textbf{Cleared} or \textbf{Tool-Infinite} catalog with \textbf{+Report} or \textbf{+Clinical Indication} planning. The orchestrator model is gpt-5.6-luna in every mode, with the same prompts. The cleared catalog is a snapshot of the FDA list used as a proxy for deployed tools; we make no regulatory claim.

\subsection{RadMatch Scoring}
\label{sec:radmatch}

Generated reports are scored against the reference report with \textbf{RadMatch}~\citep{corbiere2026}, a finding-level metric. An extractor LLM decomposes both reports into atomic findings, each with a clinical status (normal or abnormal), descriptors, measurements, a longitudinal status and a clinical-significance tier (critical, urgent, notable, routine) adapted from the ACR actionable-reporting framework. A judge LLM groups reference and generated findings that describe the same clinical entity, irrespective of wording; a status conflict (``pleural effusion'' versus ``no pleural effusion'') is recorded as a status error, and an anatomical-site mismatch as an omission on the reference side and a spurious finding on the generated side. Each matched, omitted or spurious unit inherits the highest tier among its findings. Throughout the paper the headline metric, which we call the \emph{RadMatch score}, is actionable recall: the share of reference findings at a critical, urgent or notable tier that the generated report recovers. The held-out runs also report actionable precision. The same prompts are used in every condition, and every finding, match and verdict is stored, so the scores can be audited.

\subsection{Attribution of Missed Findings, Catalog Augmentation and Orchestrator Model}
\label{sec:gap-analysis}

Every reference finding that the generated report omitted or reported incorrectly in the realistic Cleared+Clinical-Indication mode is attributed to a cause. The analysis uses only the RadMatch artifacts (the reference findings and the matching results). Misses are extracted deterministically from the matching results: every reference finding that the generated report missed or reported incorrectly is retained, with no further selection. Reports without any such finding are excluded, which leaves 9{,}675 findings across 1{,}034 of the 1{,}059 reports.

Each missed finding is adjudicated by one LLM judge (gpt-5.4-mini), which receives the full reference finding list for context, the tool catalog, and the class vocabularies of the detection, segmentation and biomarker tools filtered to the modality of the study, as the orchestrator saw them, and names the tool and class that would have been required to surface the finding. A class therefore counts as available only on the modalities it is cleared for: pleural effusion, for example, is cleared for radiography and ultrasound and is treated as absent on CT. Each judgment falls into one of five causes: (i)~\emph{not in the catalog for the modality}, no class covers the finding on the study's modality; (ii)~\emph{offered for the modality, not called}, a matching class was in the enum offered to the orchestrator and was never requested; (iii)~\emph{tool existed, not called}, a class-free tool (measurement, viewer interaction) would have surfaced the finding and was not invoked; (iv)~\emph{needs a new kind of tool}, no existing tool shape can produce the finding; (v)~\emph{no imaging tool could find it} (clinical history, demographics, acquisition technique). We call cause (i) the \emph{catalog gap} and causes (ii) and (iii) the \emph{orchestration gap}. Each judgment is grounded in the verbatim text of the finding. Running the attribution in the realistic mode, rather than in +Report where the Planner knows the answer, is what allows the two gaps to be separated.

\paragraph{Recurring missing tools.} Each label proposed under cause (i) names a class that the catalog lacks on the study's modality; since detection tools are cleared per modality, we call such a class on a modality a \emph{missing tool}. The labels are pooled across reports, with equivalent phrasings collapsed into one class, and ranked by the number of distinct reports in which they recur. A missing tool is retained at cutoff $N$ if it is missed in at least $N$ reports.

\paragraph{Catalog augmentation.} To measure what adding tools buys, the missing tools are mined from the 659 derivation reports only and added to the detection enum of their modality at three cutoffs: at least 10 reports (24 tools), at least 5 (99 tools) and at least 2 (402 tools). The detection list grows from 260, 171, 149 and 137 classes on CT, MR, radiography and ultrasound to 276, 173, 150 and 139 at the first cutoff, 317, 186, 154 and 148 at the second, and 436, 278, 174 and 180 at the third. Each augmented catalog is run in the Cleared+Clinical-Indication mode, with the same orchestrator and prompts, on 100 held-out test reports, a modality-stratified subset (CT 42, MR 26, radiography 20, ultrasound 12); the Cleared and Tool-Infinite modes are rescored on the same 100 reports with the same RadMatch version. We report actionable recall and precision, the gain over the cleared catalog, and the share of the Cleared-to-Tool-Infinite gap closed, $(\text{augmented} - \text{Cleared}) / (\text{Tool-Infinite} - \text{Cleared})$, with 95\% intervals from a paired bootstrap over reports.

\paragraph{Orchestrator model.} To measure how much of the Tool-Infinite result depends on the orchestrator, the Tool-Infinite+Clinical-Indication mode is re-run on the same 100 held-out reports with two smaller orchestrators in place of gpt-5.6-luna: gpt-5.4-nano, and Qwen3.5 9B run locally. Prompts, the tool simulator (gpt-5.6-luna) and the scorer are unchanged. A study for which an orchestrator produced no report is scored as an empty report; the score on the studies it completed is also given.

\section{Results}
\label{sec:results}

The four modes of Section~\ref{sec:modes} are run on the same 1{,}059 reports, with every tool output simulated from the reference report, so the only variables are the catalog and the information given to the Planner. Table~\ref{tab:metrics_summary} reports the RadMatch score (actionable recall, Section~\ref{sec:radmatch}).

\begin{table}[!htbp]
  \centering
  \small
  \begin{tabular}{lcc}
    \toprule
     & \multicolumn{2}{c}{\textbf{Information access to the Planner}} \\
    \cmidrule(lr){2-3}
    \textbf{Tool catalog} & +Clinical Indication (realistic) & +Report (oracle) \\
    \midrule
    Cleared (1{,}989 classes) & 0.290 & 0.426 \\
    Tool-Infinite & \textbf{0.437} & \textbf{0.740} \\
    \bottomrule
  \end{tabular}
  \caption{RadMatch score~\citep{corbiere2026} for the four evaluation modes on the same 1{,}059 PARROT reports (orchestrator gpt-5.6-luna; every tool output simulated from the reference report). The RadMatch score is actionable recall, the share of clinically actionable reference findings that the generated report recovers; higher is better, best per column in \textbf{bold}.}
  \label{tab:metrics_summary}
\end{table}

\paragraph{Tool availability.} In the realistic Clinical-Indication setting, moving from the cleared catalog to Tool-Infinite raises actionable recall from 0.290 to 0.437. With the report available to the Planner, the same change raises it from 0.426 to 0.740.

\paragraph{Information given to the Planner.} With the same tool-infinite catalog, planning from the report instead of the indication raises recall from 0.437 to 0.740; with the cleared catalog, from 0.290 to 0.426. In this setup, what the Planner knows changes recall more than what the catalog contains. Even with any tool on demand and a plan written from the report, recall stays at 0.740.

\subsection{Where the Misses Come From}
\label{sec:attribution}

Table~\ref{tab:attribution} attributes the 9{,}675 findings missed or incorrectly reported by the cleared-catalog orchestrator in the Clinical-Indication mode to the five causes of Section~\ref{sec:gap-analysis}, with each class counted as available only on the modalities it is cleared for. 61.8\% correspond to a class absent from the cleared catalog for the study's modality. 31.9\% had a cleared tool offered: in 18.3\% of cases the class was in the modality's enum and was never requested, and in 13.6\% a class-free tool would have surfaced the finding. New tool types account for 6.1\% and non-imaging findings for 0.2\%. By modality, the catalog and orchestration shares are 59\% and 35\% for CT, 68\% and 27\% for MR, 54\% and 38\% for radiography, and 64\% and 24\% for ultrasound.

\begin{table}[!htbp]
  \centering
  \footnotesize
  \setlength{\tabcolsep}{4pt}
  \begin{tabular}{@{}p{4.6cm}p{4.3cm}rr@{}}
    \toprule
    \textbf{Cause} & & \textbf{Findings} & \textbf{Share} \\
    \midrule
    \multicolumn{4}{l}{\textit{Catalog gap}} \\
    \quad (i) Not in the catalog for the modality & no class for the finding on the study's modality & 5{,}981 & 61.8\% \\
    \multicolumn{4}{l}{\textit{Orchestration gap}} \\
    \quad (ii) Offered for the modality, not called & class in the modality's enum, never requested & 1{,}772 & 18.3\% \\
    \quad (iii) Tool existed, not called & class-free tool would have surfaced it & 1{,}316 & 13.6\% \\
    \multicolumn{4}{l}{\textit{Neither}} \\
    \quad (iv) Needs a new kind of tool & no existing tool shape could produce it & 587 & 6.1\% \\
    \quad (v) No imaging tool could find it & history, demographics, technique & 19 & 0.2\% \\
    \midrule
    Total & & 9{,}675 & 100\% \\
    \bottomrule
  \end{tabular}
  \caption{Attribution of the findings missed or incorrectly reported by the cleared-catalog orchestrator in the realistic Cleared\,+\,Clinical-Indication mode (1{,}034 reports with at least one miss; one LLM judgment per finding). A class counts as available only on the modalities it is cleared for.}
  \label{tab:attribution}
\end{table}

The classes the orchestrator was offered and did not request are common ones: vascular stenosis (not requested in 46 reports, 157 findings), aortic calcification (46), atherosclerosis (39), pleural effusion (39, on radiography and ultrasound), brain-tissue abnormalities (31), gallbladder (29), atrophy and fracture (24 each), chronic infarct (23) and vertebral compression fracture (21). The missing tools are common findings as well: pleural effusion on CT (missing in 69 reports, 79 findings), atrophy on CT (41), adrenal gland on CT (40), mediastinal lymphadenopathy and consolidation on CT (35 each), hydronephrosis on ultrasound (29), venous thrombosis on MR (28), liver lesion on CT (26), then bile-duct dilatation (24), atelectasis (21), lymphadenopathy (20), hilar lymphadenopathy and pleural thickening (19 each), renal cyst (17) and spondylosis (15), all on CT. Of the 3{,}088 distinct missing tools proposed under cause (i), 96.9\% of the findings map to \texttt{ai\_detect\_anomalies}; 2{,}441 are seen in a single report, while the 165 proposed in at least five reports cover 34\% of the catalog gap (2{,}024 of its 5{,}981 findings) and the 50 proposed in at least ten cover 19\%.

\subsection{Adding the Missing Tools}
\label{sec:augmentation}

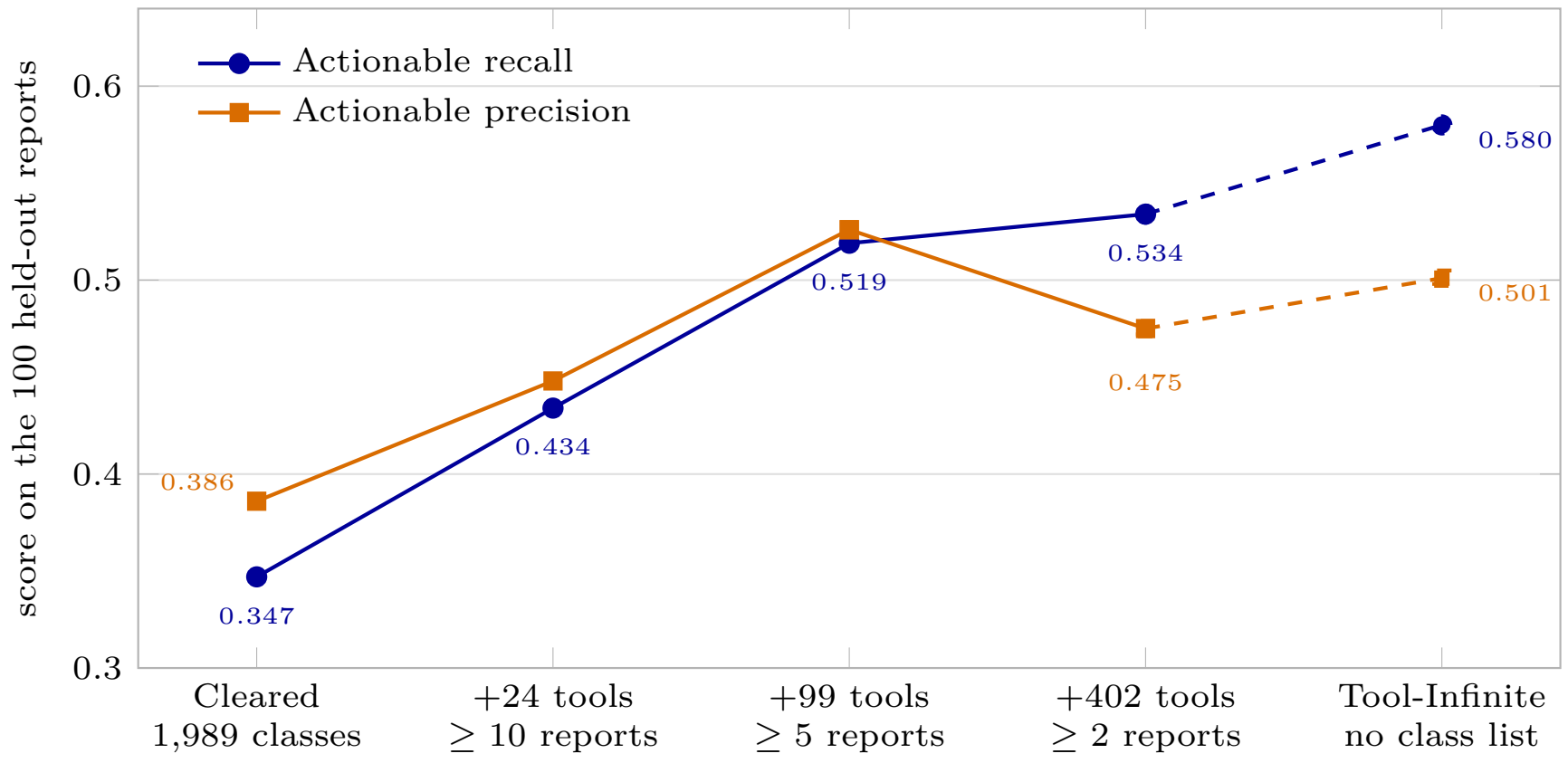
\begin{figure}[!htbp]
\centering
\begin{tikzpicture}
\begin{axis}[
  width=0.82\linewidth, height=6.4cm,
  xmin=0.6, xmax=5.4, ymin=0.30, ymax=0.64,
  xtick={1,2,3,4,5},
  xticklabels={{Cleared\\1{,}989 classes},{$+24$ tools\\$\geq 10$ reports},{$+99$ tools\\$\geq 5$ reports},{$+402$ tools\\$\geq 2$ reports},{Tool-Infinite\\no class list}},
  x tick label style={font=\scriptsize, align=center},
  y tick label style={font=\scriptsize},
  ylabel={score on the 100 held-out reports}, ylabel style={font=\scriptsize},
  ymajorgrids, grid style={gray!25},
  axis line style={gray!60}, tick style={gray!60},
  legend style={at={(0.03,0.97)}, anchor=north west, font=\scriptsize, draw=none, fill=none}, legend cell align=left,
  nodes near coords, nodes near coords style={font=\tiny, anchor=north, yshift=-3pt},
  every node near coord/.append style={/pgf/number format/fixed, /pgf/number format/precision=3, /pgf/number format/fixed zerofill},
]
\addplot[blue!60!black, thick, mark=*, mark size=1.8pt] coordinates {(1,0.347) (2,0.434) (3,0.519) (4,0.534)};
\addplot[blue!60!black, thick, dashed, mark=*, mark size=1.8pt, forget plot, nodes near coords style={font=\tiny, anchor=west, xshift=4pt}, point meta=explicit symbolic] coordinates {(4,0.534) [] (5,0.580) [0.580]};
\addplot[orange!85!black, thick, mark=square*, mark size=1.6pt, nodes near coords style={font=\tiny, anchor=south east, xshift=-1pt, yshift=2pt}, point meta=explicit symbolic] coordinates {(1,0.386) [0.386] (2,0.448) [] (3,0.526) []};
\addplot[orange!85!black, thick, mark=square*, mark size=1.6pt, forget plot, nodes near coords style={font=\tiny, anchor=north, yshift=-3pt}, point meta=explicit symbolic] coordinates {(3,0.526) [] (4,0.475) [0.475]};
\addplot[orange!85!black, thick, dashed, mark=square*, mark size=1.6pt, forget plot, nodes near coords style={font=\tiny, anchor=west, xshift=4pt}, point meta=explicit symbolic] coordinates {(4,0.475) [] (5,0.501) [0.501]};
\legend{Actionable recall, Actionable precision}
\end{axis}
\end{tikzpicture}
\caption{Actionable recall and precision as the cleared catalog is augmented with the missing tools (classes absent on a modality) missed in at least $N$ of the 659 derivation reports, scored on the same 100 held-out reports at every step. The dashed segment leads to the Tool-Infinite mode, where the orchestrator can request any class. All values and the 95\% intervals on the paired gains are given in Table~\ref{tab:gap}.}
\label{fig:gap}
\end{figure}

Table~\ref{tab:gap} and Figure~\ref{fig:gap} show the effect of adding the recurring missing tools. On the 100 held-out reports the cleared catalog scores 0.347 actionable recall (167 of 481 reference findings recovered) and Tool-Infinite 0.580 (282 of 486), a gap of 0.23. Adding the 24 tools missed in at least 10 derivation reports brings recall to 0.434 ($+0.09$, 95\% CI $+0.02$ to $+0.16$), 37\% of the gap (CI 8 to 65\%). Adding the 99 tools at the $\geq 5$ cutoff brings it to 0.519 ($+0.17$, CI $+0.11$ to $+0.24$), 74\% of the gap (CI 51 to 105\%). The 402 tools at $\geq 2$ bring it to 0.534 ($+0.19$, CI $+0.10$ to $+0.27$; 80\% of the gap), no clear gain over $\geq 5$ for four times as many tools: the extra 303 draw calls (1{,}237 calls to added tools, against 615 at $\geq 5$) with little change in what is recovered. Actionable precision rises with the first tools and falls back at $\geq 2$ (0.386 with the cleared catalog, 0.526 at $\geq 5$, 0.475 at $\geq 2$); Tool-Infinite scores 0.501.

\begin{table}[!htbp]
  \centering
  \scriptsize
  \setlength{\tabcolsep}{3pt}
  \begin{tabular}{@{}lrccccr@{}}
    \toprule
    & \textbf{Tools added} & & & \multicolumn{2}{c}{\textbf{vs.\ Cleared [95\% CI]}} & \\
    \cmidrule(lr){2-2} \cmidrule(lr){5-6}
    \textbf{Catalog} & & \textbf{Recall} & \textbf{Prec.} & Gain & Gap closed & Calls (pos.) \\
    \midrule
    Cleared              & 0   & 0.347 & 0.386 & ---                  & ---            & --- \\
    $+$ tools $\geq 10$  & 24  & 0.434 & 0.448 & $+0.09$ [0.02,\,0.16] & 37\% [8,\,65]   & 217\,(45) \\
    $+$ tools $\geq 5$   & 99  & 0.519 & 0.526 & $+0.17$ [0.11,\,0.24] & 74\% [51,\,105] & 615\,(123) \\
    $+$ tools $\geq 2$   & 402 & 0.534 & 0.475 & $+0.19$ [0.10,\,0.27] & 80\% [52,\,110] & 1{,}237\,(243) \\
    Tool-Infinite        & --- & 0.580 & 0.501 & $+0.23$              & 100\%          & --- \\
    \bottomrule
  \end{tabular}
  \caption{Adding the missing tools. The cleared catalog is augmented, in the realistic Cleared\,+\,Clinical-Indication mode, with the missing tools (classes absent from the catalog on a modality) missed in at least 10, 5 or 2 of the 659 derivation reports (the resulting detection-list sizes per modality are given in Section~\ref{sec:gap-analysis}). All five rows are scored on the same 100 held-out test reports (CT 42, MR 26, radiography 20, ultrasound 12); RadMatch actionable recall and precision, higher is better. Gap closed $= (\text{augmented} - \text{Cleared}) / (\text{Tool-Infinite} - \text{Cleared})$; intervals are 95\% paired bootstraps over reports. \emph{Calls}: number of tool calls that requested an added tool, with the number that returned a positive result in parentheses. Reference findings are re-extracted at each scoring, so denominators differ slightly across rows (464 to 486).}
  \label{tab:gap}
\end{table}

\begin{figure}[p]
\centering
\includegraphics[width=\linewidth]{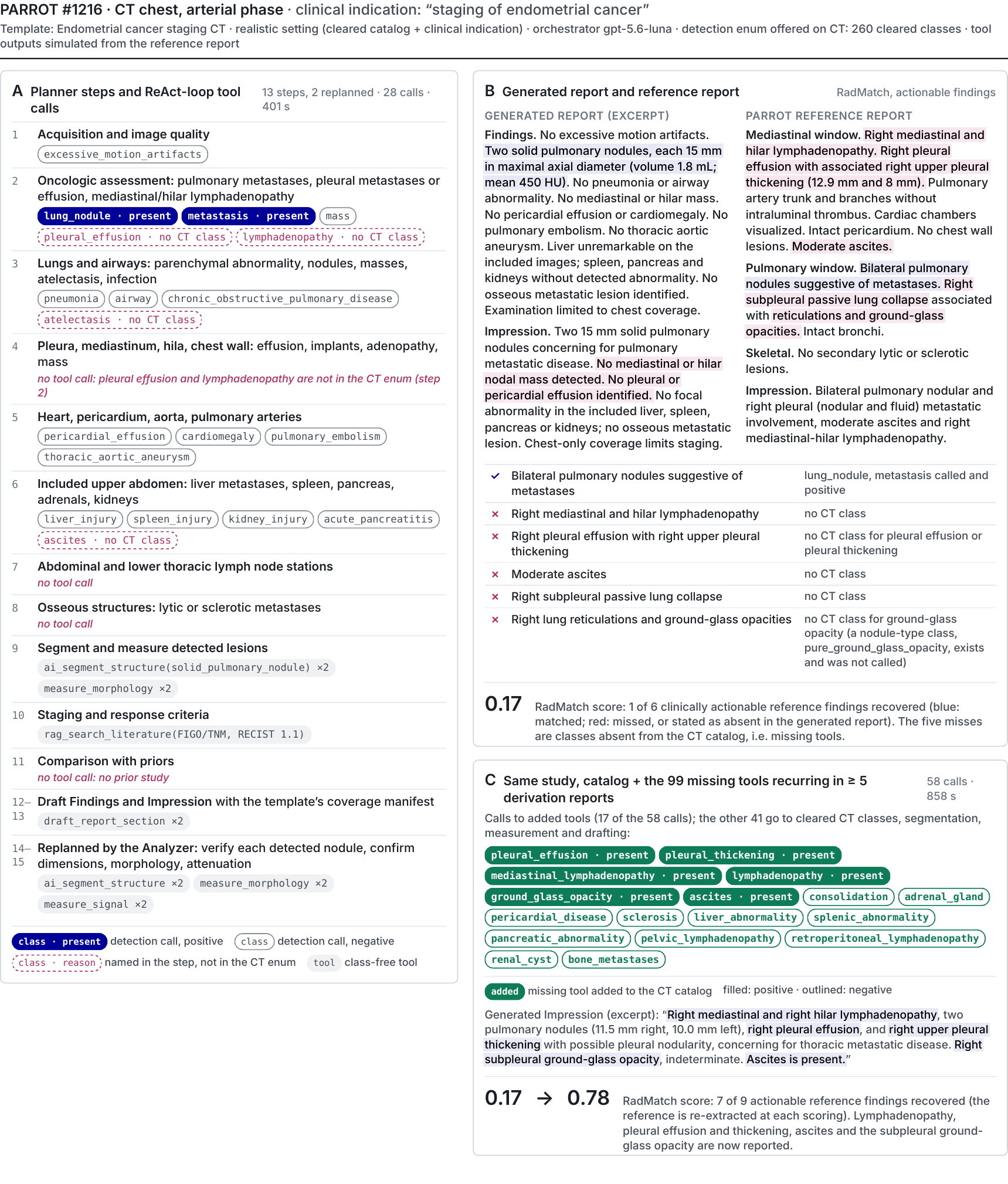}
\caption{Worked case study (PARROT \#1216, chest CT for endometrial cancer staging) in the realistic Cleared\,+\,Clinical-Indication mode. (\textbf{A}) The Planner's steps and the tool calls the ReAct loop made for each, with the classes requested from the 260-class CT detection enum and the simulator's answer; dashed chips are findings the step names for which the enum has no class, and three steps end without a call. (\textbf{B}) The generated report against the PARROT reference and the actionable reference findings matched by RadMatch: the pulmonary nodules are recovered; lymphadenopathy, pleural effusion and thickening, ascites and the subpleural changes are not, two of them being stated as absent. (\textbf{C}) The same study with the 99 missing tools of Section~\ref{sec:augmentation} added to the catalog: six positive calls to added tools bring the RadMatch score from 0.17 to 0.78.}
\label{fig:evaluation}
\end{figure}

The 99 tools are mostly common findings, and the 615 calls to them at $\geq 5$ returned 123 positive results, led by atelectasis, disc bulge, degenerative spine changes, lymphadenopathy, spondylosis, brain-tissue abnormalities, osteophytes, osteoarthritis, edema and pleural effusion on CT. Figure~\ref{fig:evaluation} follows one held-out study through the realistic mode, before and after the 99 tools are added.

\subsection{Changing the Orchestrator}
\label{sec:orchestrator}

\begin{figure}[!htbp]
\centering
\begin{tikzpicture}
\begin{axis}[
  ybar, scale only axis, width=6.2cm, height=4.6cm, bar width=16pt,
  ymin=0, ymax=0.8, xmin=0.4, xmax=3.6,
  xtick={1,2,3}, xticklabels={{gpt-5.6-luna},{gpt-5.4-nano},{Qwen3.5 9B\\(local)}},
  ytick={0,0.2,0.4,0.6},
  x tick label style={font=\scriptsize, align=center}, y tick label style={font=\scriptsize},
  ylabel={RadMatch score, 100 held-out reports}, ylabel style={font=\scriptsize},
  ymajorgrids, grid style={gray!25},
  axis line style={gray!60}, tick style={gray!60},
  nodes near coords, nodes near coords style={font=\tiny, anchor=south, yshift=12pt},
  every node near coord/.append style={/pgf/number format/fixed, /pgf/number format/precision=3, /pgf/number format/fixed zerofill},
  legend style={at={(0.97,0.86)}, anchor=north east, font=\scriptsize, draw=none, fill=none}, legend cell align=left,
]
\addplot[fill=blue!60!black, draw=none, forget plot, error bars/.cd, y dir=both, y explicit, error bar style={gray!70, thin}, error mark options={gray!70, rotate=90, mark size=1.5pt}] table[x=x, y=r, y error plus=ep, y error minus=em] {
x r ep em
1 0.580 0.066 0.066
2 0.355 0.072 0.066
3 0.356 0 0
};
\draw[gray!60!black, dashed, thick] (axis cs:0.4,0.347) -- (axis cs:3.6,0.347);
\addlegendimage{line legend, gray!60!black, dashed, thick}
\addlegendentry{luna, cleared catalog: 0.347}
\end{axis}
\end{tikzpicture}
\caption{Changing the orchestrator. RadMatch score (actionable recall, with 95\% bootstrap interval) in the Tool-Infinite\,+\,Clinical-Indication mode on the same 100 held-out reports, with gpt-5.6-luna, gpt-5.4-nano and Qwen3.5 9B (run locally) as orchestrator; prompts, tool simulator and scorer are the same in every run. Qwen's 10 failed studies are scored as empty reports. The dashed line is luna restricted to the cleared catalog on the same reports.}
\label{fig:orchestrator}
\end{figure}
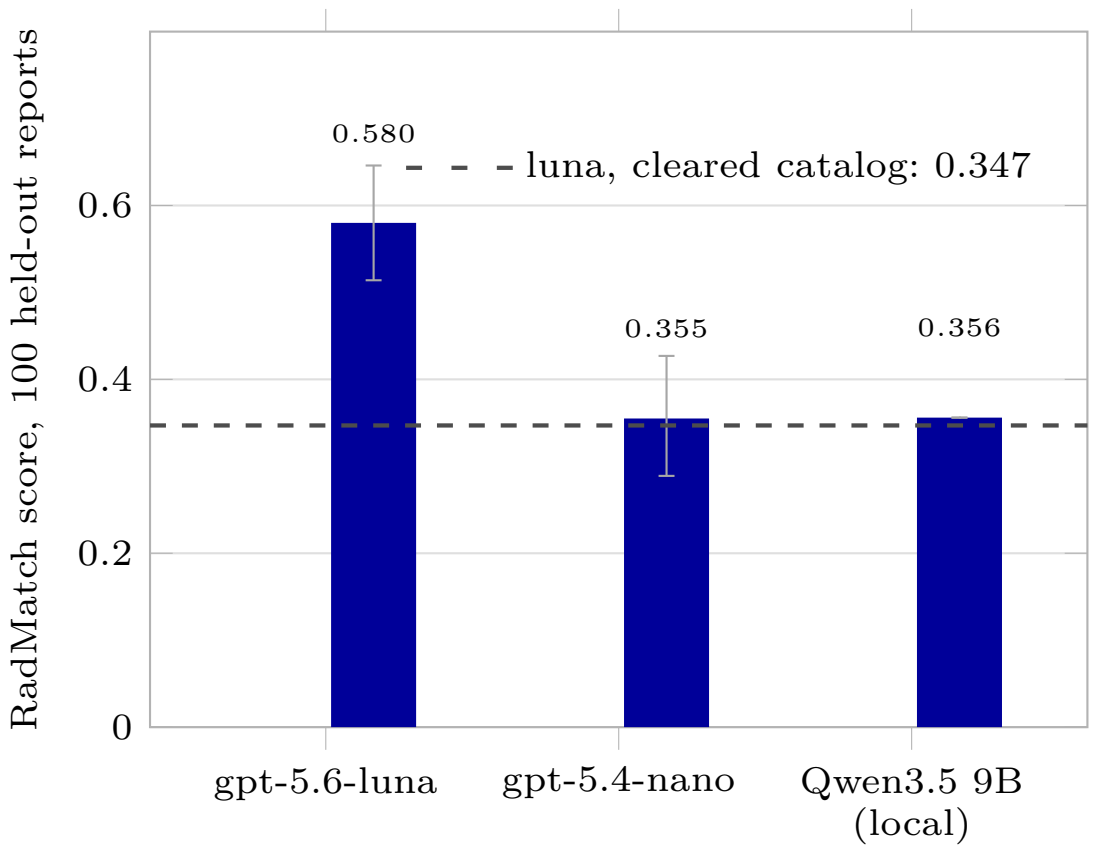

Figure~\ref{fig:orchestrator} compares the three orchestrators on the same 100 held-out reports in the Tool-Infinite+\allowbreak Clinical-Indication mode. With any class available, gpt-5.4-nano scores 0.355 (174 of 490 actionable findings), 0.225 below gpt-5.6-luna (95\% CI $-0.30$ to $-0.15$) and at the level of luna restricted to the cleared catalog (0.347): its gain over that run is $+0.01$ (CI $-0.08$ to $+0.10$). Qwen3.5 9B, run locally, produced no report for 10 of the 100 studies (4 failed at planning, 6 returned an empty report); scoring those as empty, it reaches 0.356 (162 of 455), $-0.22$ against luna (CI $-0.29$ to $-0.16$), and 0.413 against 0.600 for luna on the 90 studies it completed ($-0.19$, CI $-0.26$ to $-0.12$). The orchestrator thus matters as much as the catalog: with every class available, the two smaller models score where luna scores with the cleared catalog, so the headroom that a richer catalog creates is only realized by an orchestrator able to use it. The converse is the promising side of the result: the Tool-Infinite ceiling is set with today's orchestrator, and future orchestrators may raise it.

\section{Discussion}

\subsection{The Catalog Gap}

Six in ten misses are classes that no cleared device offers on the study's modality, and the recurring ones are ordinary: pleural effusion on CT, mediastinal lymphadenopathy, atrophy, bile-duct dilatation, atelectasis, ascites. Adding tools for these recurring findings helps: on held-out reports, 99 tools close most of the gap to the tool-infinite catalog and the next 303 add little. Two caveats: the sample is 100 reports, with an interval of 51\% to 105\% on the share of the gap closed; and the ceiling is the tool-infinite level, 0.437 actionable recall on the full corpus. Above all, every tool here is perfect: the simulator answers from the reference report, so an added tool never misses, never raises a false alarm and never disagrees with the radiologist, which is far from the clinical situation. The gains measure what the catalog can offer, not what deployed tools would deliver. The catalog is also thin: 59\% of its classes rest on a single device, and only 29\% of FDA-authorized radiology AI devices reported clinical testing in a recent review~\citep{sivakumar2025}, as in earlier surveys of commercial products~\citep{vanleeuwen2021,vanleeuwen2025}.

\subsection{The Orchestration Gap}

In 32\% of the misses a cleared tool was offered for the study's modality and not called: vascular stenosis and aortic calcification in 46 reports each, pleural effusion in 39 on radiography and ultrasound. More classes do not help here. The attribution does not say why, but the uncalled concepts (stenosis, calcification, atherosclerosis, effusion, atrophy) are ones a radiologist checks on most studies whatever the indication, so the Planner's coverage manifest does not yet produce a systematic survey. Two other results point the same way: with the same unlimited catalog, planning from the report instead of the indication raises recall from 0.437 to 0.740; and with every class available, gpt-5.4-nano and Qwen3.5 9B score where luna scores with the cleared catalog, so that on the same 100 reports the catalog change is worth $+0.23$ with luna and $+0.01$ with nano. The orchestrator decides how much a catalog is worth. Skipping an available tool is a documented failure mode of LLM agents outside radiology~\citep{soni2026toolfailbench}; in clinical benchmarks, tools have not reliably improved accuracy~\citep{liu2026agents}, and radiology agent benchmarks report execution success well above task success~\citep{llmagentcores,maksudov2026abra}.

\subsection{The Residual Ceiling}

With any tool on demand, perfect simulated outputs and a plan written from the report, actionable recall is 0.740: a quarter of the actionable findings stays out of reach of tool composition. The remainder includes findings that only make sense in combination, lesions that need characterization in context, and descriptions that do not decompose into tool calls; part of it is by construction, since no image-to-text VLM is in the pipeline and whatever cannot be phrased as a tool call is unreachable. A deployed system could instead delegate such steps to a VLM; this fallback is disabled in all experiments so that missing tools remain visible. Radiology is not solved by composing tools, and this remainder is where future orchestrators, and models that read the images themselves, have the most to gain.

\subsection{Limitations}

PARROT reports are fictional, which gives scale and privacy but is not a proxy for real-world clinical situations~\citep{parrot}. Evaluation is text-only and every tool call is simulated from the reference report, so tool error is excluded by design and broad classes return more than real detectors would. The held-out runs use 100 reports, enough for the direction of the effect but not to rank the cutoffs. The catalog is a snapshot (FDA decisions to March 2026) built with LLM extraction and consolidation, and the main orchestrator is one model; absolute numbers will move with the model.

\section{Conclusion}

Agentic systems can already compose today's cleared radiology tools into a report, and this paper measures how far that composition goes. Through a shared interface, an orchestrator planning from the clinical indication recovers close to a third of the actionable findings with the cleared catalog, and more than twice as many when any tool can be requested, with a plan matching the report. Most of the distance between the two is coverage, and adding tools for the recurring missing findings, nearly all of them common, would reduce it; on held-out reports a few dozen such tools close most of that distance, with perfect tool outputs that real tools will not match. The rest is orchestration: even with perfect tools and a perfect plan, a quarter of the actionable findings stays out of reach, underlining the challenges ahead for the next generation of agentic systems.

\section*{Acknowledgements}

We thank Jun Ma (University Health Network, Toronto, Canada) for his careful reading of the manuscript and his comments.

\section*{Use of AI}

Claude (Anthropic) was used to compose the manuscript and some of the figure panels. The experiments were designed by the authors, and all data, text, figures and numbers were reviewed and verified by them.

\bibliographystyle{splncs04}
\bibliography{references}

\begin{thebibliography}{10}
\providecommand{\url}[1]{\texttt{#1}}
\providecommand{\urlprefix}{URL }
\providecommand{\doi}[1]{https://doi.org/#1}

\bibitem{mcp}
{Anthropic}: Model context protocol specification. \url{https://modelcontextprotocol.io} (2024)

\bibitem{bluethgen2025}
Bluethgen, C., Van~Veen, D., Truhn, D., Kather, J.N., Moor, M., Polacin, M., Chaudhari, A., Frauenfelder, T., Langlotz, C.P., Krauthammer, M., Nooralahzadeh, F.: Agentic systems in radiology: Design, applications, evaluation, and challenges. arXiv preprint arXiv:2510.09404  (2025)

\bibitem{petctagent}
Choi, H., Bae, S., Na, K.J.: End-to-end {PET/CT} interpretation and quantification with an {LLM}-orchestrated {AI} agent: A real-world pilot study. medRxiv  (2026), doi:10.64898/2026.02.21.26346798

\bibitem{corbiere2026}
Corbi\`ere, C., Machado, L., Charley, A., Callard, B., Manceron, P., Dancette, C.: {RadMatch}: Auditable radiology report evaluation via finding-level matching. arXiv preprint arXiv:2609.01470  (2026)

\bibitem{medrax}
Fallahpour, A., et~al.: {MedRAX}: Medical reasoning agent for chest {X}-ray. arXiv preprint arXiv:2502.02673  (2025)

\bibitem{ctflow}
Gu, Y., Zhang, X., Mu, L., Yu, Y., Huang, Z., Zhang, S., Zhang, X.: {CT}-flow: Orchestrating {CT} interpretation workflow with model context protocol servers. arXiv preprint arXiv:2603.00123  (2026)

\bibitem{ct2rep}
Hamamci, I.E., Er, S., Menze, B.: {CT2Rep}: Automated radiology report generation for {3D} medical imaging. arXiv preprint arXiv:2403.06801  (2024)

\bibitem{jiang2026radonc}
Jiang, C., Wei, S., Zhao, S., Li, J., Chen, J., Liu, W.: {RadOnc}-agent: An {LLM}-orchestrated framework for {AI} workflows across the radiotherapy care pathway. arXiv preprint arXiv:2610.06923  (2026)

\bibitem{khandelwal}
Khandelwal, A., Matta, R.: Agentic {AI} in radiology: Emerging potential and unresolved challenges. British Journal of Radiology  (2025)

\bibitem{parrot}
Le~Guellec, B., Adambounou, K., et~al.: {PARROT}: An open multilingual radiology reports dataset. arXiv preprint arXiv:2507.22939  (2025)

\bibitem{vanleeuwen2021}
van Leeuwen, K.G., Schalekamp, S., Rutten, M.J.C.M., van Ginneken, B., de~Rooij, M.: Artificial intelligence in radiology: 100 commercially available products and their scientific evidence. European Radiology  \textbf{31}(6),  3797--3804 (2021)

\bibitem{vanleeuwen2025}
van Leeuwen, K.G., et~al.: Artificial intelligence in radiology: 173 commercially available products and their scientific evidence. European Radiology  (2025), doi:10.1007/s00330-025-11830-8

\bibitem{liu2026agents}
Liu, Y., Carrero, Z.I., Jiang, X., Ferber, D., W{\"o}lflein, G., Zhang, L., Jayabalan, S., Lenz, T., Hui, Z., et~al.: Benchmarking large language model-based agent systems for clinical decision tasks. npj Digital Medicine  \textbf{9}, ~259 (2026), doi:10.1038/s41746-026-02443-6

\bibitem{maksudov2026abra}
Maksudov, B., Kurenkov, V., Curran, K.M., Mileo, A.: {ABRA}: Agent benchmark for radiology applications. arXiv preprint arXiv:2605.11224  (2026)

\bibitem{templatecollapse}
Maye-Lasserre, T., Li, Y., Jian, B., Ghahremani, M., Wiestler, B., Wachinger, C.: Generating reports or repeating templates? measuring and mitigating template collapse in {3D} {CT} report generation. arXiv preprint arXiv:2605.30984  (2026)

\bibitem{radreport}
{Radiological Society of North America}: {RadReport.org}: Radiology reporting templates. \url{https://radreport.org} (2026), accessed October 2026

\bibitem{sivakumar2025}
Sivakumar, R., Lue, B., Kundu, S.: {FDA} approval of artificial intelligence and machine learning devices in radiology: A systematic review. JAMA Network Open  \textbf{8}(11),  e2542338 (2025), doi:10.1001/jamanetworkopen.2025.42338

\bibitem{soni2026toolfailbench}
Soni, H.: {ToolFailBench}: Diagnosing tool-use failures in {LLM} agents. arXiv preprint arXiv:2607.04686  (2026)

\bibitem{tripathi}
Tripathi, S., et~al.: Agentic {AI} in radiology. Radiology  (2026)

\bibitem{fda-aiml}
{U.S. Food and Drug Administration}: Artificial intelligence and machine learning ({AI/ML})-enabled medical devices. \url{https://www.fda.gov/medical-devices/software-medical-device-samd/artificial-intelligence-and-machine-learning-aiml-enabled-medical-devices} (2025), accessed 2025

\bibitem{wang2026hallucxr}
Wang, H., Li, Z.: {HalluCXR}: Benchmarking and mitigating hallucinations in medical vision-language models for chest radiograph interpretation. arXiv preprint arXiv:2605.20469  (2026)

\bibitem{medagent3d}
Wang, Z., Cai, L., et~al.: {3DMedAgent}: Unified perception-to-understanding for {3D} medical analysis. arXiv preprint arXiv:2602.18064  (2026)

\bibitem{react}
Yao, S., Zhao, J., Yu, D., Du, N., Shafran, I., Narasimhan, K., Cao, Y.: {ReAct}: Synergizing reasoning and acting in language models. arXiv preprint arXiv:2210.03629  (2022)

\bibitem{radiologistcopilot}
Yu, S., et~al.: Radiologist copilot: An {LLM}-based agentic {AI} framework for radiology reporting. arXiv preprint arXiv:2512.02814  (2025)

\bibitem{llmagentcores}
Zheng, Q., Wu, C., Zhao, W., Qiu, P., Dai, L., Zhang, Y., Wang, Y., Xie, W.: How well can modern {LLMs} act as agent cores in radiology environments? arXiv preprint arXiv:2412.09529  (2024)

\end{thebibliography}

\end{document}